\documentclass[11pt]{article}

\usepackage[]{acl}
\usepackage{color}
\usepackage{listings}
\usepackage{xcolor}

\usepackage{times}
\usepackage{amsmath}
\usepackage{latexsym}
\usepackage{graphicx}
\usepackage{subfigure}

\usepackage[utf8]{inputenc}

\usepackage{microtype}

\usepackage{inconsolata}

\title{SCCA: Shifted Cross Chunk Attention for long contextual semantic expansion}

\author{Yuxiang Guo \\
  Beihang University  \\
  \texttt{irisg@buaa.edu.cn} 
  }
\begin{document}
\maketitle
\begin{abstract}
Sparse attention as a efficient method can significantly decrease the computation cost, but current sparse attention tend to rely on window self attention which block the global information flow. For this problem, we present Shifted Cross Chunk Attention (SCCA), using different KV shifting strategy to extend respective field in each attention layer. Except, we combine Dilated Attention(DA) and Dilated Neighborhood Attention(DNA) to present Shifted Dilated Attention(SDA). Both SCCA and SDA can accumulate attention results in multi head attention to obtain approximate respective field in full attention. In this paper, we conduct language modeling experiments using different pattern of SCCA and combination of SCCA and SDA. The proposed shifted cross chunk attention (SCCA) can effectively extend large language models (LLMs) to longer context combined with Positional interpolation(PI) and LoRA than current sparse attention.. Notably, SCCA adopts LLaMA2 7B from 4k context to 8k in single V100. This attention pattern can provide a Plug-and-play fine-tuning method to extend models’ context while retaining their original architectures, and is compatible with most existing techniques, like FlashAttention-2.
\end{abstract}

\section{Introduction}
The Transformer architecture is rapidly becoming one of the most widely applied deep learning architectures, and the emergence of Large Language Models (LLMs) using Transformer has brought improvements to many tasks. However, a significant challenge lies in the quadratic computation complexity introduced by the vanilla transformer, which hinders the increase in input length.

Some researchers opt for using sparse attention patterns to reduce computing complexity and save memory. While sparse transformers like local attention \cite{qiu-etal-2020-blockwise} and sliding window context \cite{beltagy2020longformer} based on window size are proposed, these attention pattern face a limitation in information flow within the window or chunk. Other approaches, such as dilated window attention \cite{beltagy2020longformer} and sparse Transformer \cite{child2019generating}, require changes to the model structure and lack a corresponding CUDA-friendly implementation. Swin Transformer \cite{Liu2021SwinTH} and Dilated Neighborhood Attention (DNA) provide a cross-layer attention pattern in chunk-based attention, introducing information flow between different chunks or windows. However, global information flow remains lacking in these methods.

While current LLMs have revolutionized language modeling and showcased impressive task performance \cite{qspaper,cohan-etal-2018-discourse,kocisky-etal-2018-narrativeqa,shi-etal-2023-effidit,huang-etal-2021-efficient,shaham-etal-2022-scrolls,bai2023longbench}, they are constrained by pre-defined context window size. The performance significantly declines when input tokens exceed these limited context length. Direct context extrapolation in LLMs using positional embedding, such as RoPE, can lead to catastrophic consequences. To address this out-of-distribution problem, various Position Interpolation algorithms \cite{chen2023extending,ntkaware,peng2023yarn} have been introduced. While these methods effectively extrapolate the length of LLMs using RoPE, full fine-tuning is still required.

Longlora \cite{chen2023longlora} introduces a new insight that sparse attention can be used in the fine-tuning process to extrapolate the context length of LLMs, resulting in non-trivial computation savings with performance similar to full fine-tuning. However, the lack of information flow persists in the length-extending process, emphasizing the importance of an information-efficient attention pattern.

In this paper, we propose Shifted Cross Chunk Attention (SCCA), which utilizes different key-value (KV) shift strategies to enable queries to directly attend outside the same window. We provide two shifting strategies, $SCCA_{fixed}$ and $SCCA_{flow}$, to introduce different information flows, with $SCCA_{flow}$ achieving approximate full attention within its respective field during the accumulation of different head attention results with linear complexity. Additionally, we combine Dilated Attention (DA) and Dilated Neighborhood Attention (DNA) to present Shifted Dilated Attention (SDA). Both SCCA and SDA can accumulate attention results in multi-head attention to obtain an approximate respective field in full attention. To evaluate the effectiveness of the attention pattern in extending LLMs' context length, we conduct language modeling experiments using different SCCA patterns and a combination of SCCA and SDA on the PG19 validation split and Proof test split. The proposed SCCA can extend LLMs to a longer context in a more efficient way, combined with Positional Interpolation (PI) and LoRA, compared to $S^2$ attention used in Longlora \cite{chen2023longlora}. Both SCCA and SDA are plug-and-play fine-tuning methods which can extend model contexts while retaining their original architectures.
\section{Related work}
\subsection{Sparse attention}
Sparse attention tend to conduct self attention operation in a sub token sets of a sequence to decrease computing time and memory. Blockwise attention, also named local attention \cite{qiu-etal-2020-blockwise} break a sequence with N tokens into n non-overlapping windows with $N/n$ tokens. The local attention allows one query to attend to tokens within the same window. Based on this window context, different sparse pattern are proposed. Sliding window attention \cite{beltagy2020longformer} adapt sliding window to conduct local attention. Dilated sliding window further increases the receptive field in a “dilated” sliding window way \cite{beltagy2020longformer}. This is analogous to dilated CNNs \cite{oord2016wavenet} where the window has gaps of size dilation d. The fixed pattern of sparse Transformer\cite{child2019generating} is composed of a local attention and a strided attention. Stried attention allow query Q attend to tokens that are not in the same window. Swin transformer\cite{Liu2021SwinTH} provide a shifted window attention to allow self-attention computation both to non-overlapping local windows and cross-window connection. similar to dilated window attention, LongNet \cite{ding2023longnet} and Dilated Neighborhood Attention\cite{hassani2023dilated} extend different window size and adapt different gaps of size dilation d.
\begin{table*}
\centering 
\begin{tabular}{ll}
\hline
Pseudocode of $SCCA_{flow}$ in PyTorch-like style.\\
\hline
\textcolor{lightgray}{\# B: batch size; H: head number; N: sequence length; D: dimension of each attention head }\\
\textcolor{lightgray}{\# index : number of heads conduct same shift pattern} \\
\textcolor{lightgray}{\# w: group size; \# H: number of attention heads;} \\
\textcolor{lightgray}{\# K and V in shape (B, H, N, D)} \\
\textcolor{lightgray}{\# Key line 2: each index heads shift KV i*w on N length sequence}\\
 for i in range(num\_group): \\
 \ \ \ \ kv[:, i*index:(i+1)*index] = qkv[:, i*index:(i+1)*index].roll(w*i, dims=2)  \\  
kv=kv.reshape(B,H,$\frac{N}{w}$,w,D) \\ \hline \\
After shifting KV we need split shifted KV into $\frac{N}{w}$ chunks \\ 
 \hline
\label{tab:shiftcode}
\end{tabular}
\end{table*}   
\subsection{Length extrapolation in LLMs}
Length extrapolation aims to ensure that the model continues to perform well, even when the number of input tokens during inference exceeds the size of the context window on which the model is trained (Press et al., 2021). While certain techniques such as ALiBi \cite{alibi} and LeX \cite{sun-etal-2023-length} enable length extrapolation of Transformers, i.e. train on short context windows and inference on longer ones, many existing pre-trained LLMs, including LLaMA (Touvron et al., 2023), use positional encodings that have weak extrapolation properties (e.g., RoPE (Su et al., 2021)). One  one question exists in these LLMs is directly extrapolate context length in inference processing can bring a catastrophic performance and training LLMs with long context from scratch is prohibitively expensive for most researchers. Position Interpolation \cite{chen2023extending} introduces a modification upon RoPE and extends the context length of LLaMA to 32768. Subsequently, a range of Positional Interpolation (PI) strategies like NTK (\cite{ntkaware}) and YaRN \cite{peng2023yarn} have been introduced. While these methods make the length extrapolation of LLMs using RoPE effective, full fine-tuning is still required. Longlora \cite{chen2023longlora} propose a new insight that sparse attention can be used in fine-tuing process to extrapolate the contex, leading to non-trivial computation saving with similar performance to fine-tuning. Different from training in a full-length, some researchers choice to design suitable training strategy to extend context length in original context window. PoSE \cite{zhu2023pose} manages to decouple train / target length, requiring only the original context size for fine-tuning.
\subsection{LongBench}
The field of NLP has long sought to endow machines with the ability to understand and reason over a long context \cite{qspaper}. Tasks such as summarization \cite{cohan-etal-2018-discourse} and question answering \cite{kocisky-etal-2018-narrativeqa} based on books, report \cite{huang-etal-2021-efficient}, and documents \cite{pang-etal-2022-quality}, and code generation at the repository level demand the ability to model long context sequences that span thousands or even tens of thousands of tokens in length \cite{scrolls}. LongBench is the first bilingual, multi-task benchmark tailored for long context understanding. LongBench \cite{bai2023longbench} is composed of 6 major task categories and 21 different tasks, covering key long-text application scenarios including multi-document QA, single-document QA, summarization, few-shot learning, code completion, and synthesis tasks. LongBench contains 4,750 test instances, with an average length of 6,711 words for English instances (including code).
\begin{figure*}
    \centering
    \subfigure[Shift half chunk]{
     \includegraphics[width=0.42\textwidth]{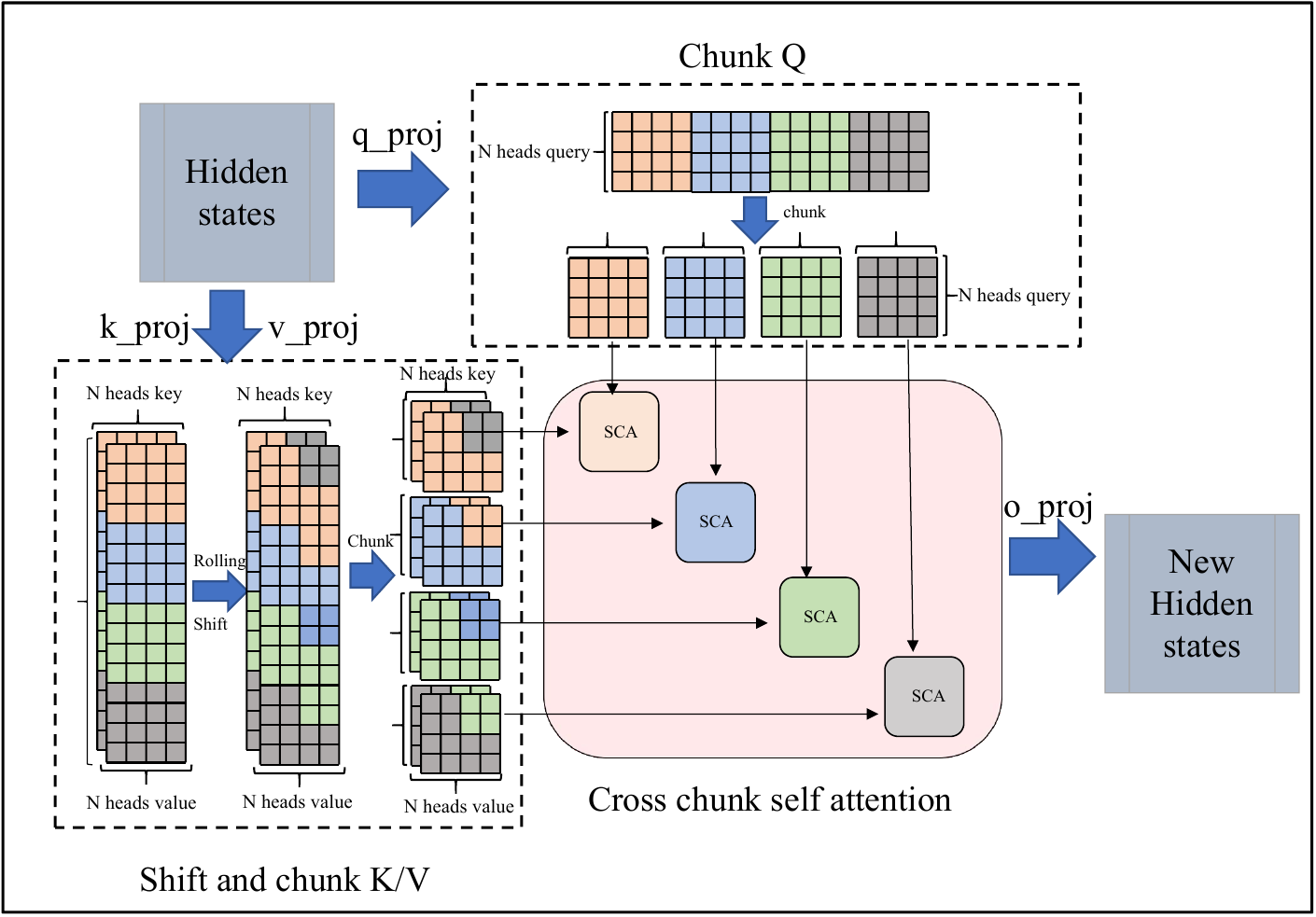}
     \label{fig:shifthalf}
    }
    \subfigure[Shift all group]{
       \includegraphics[width=0.42\textwidth ]{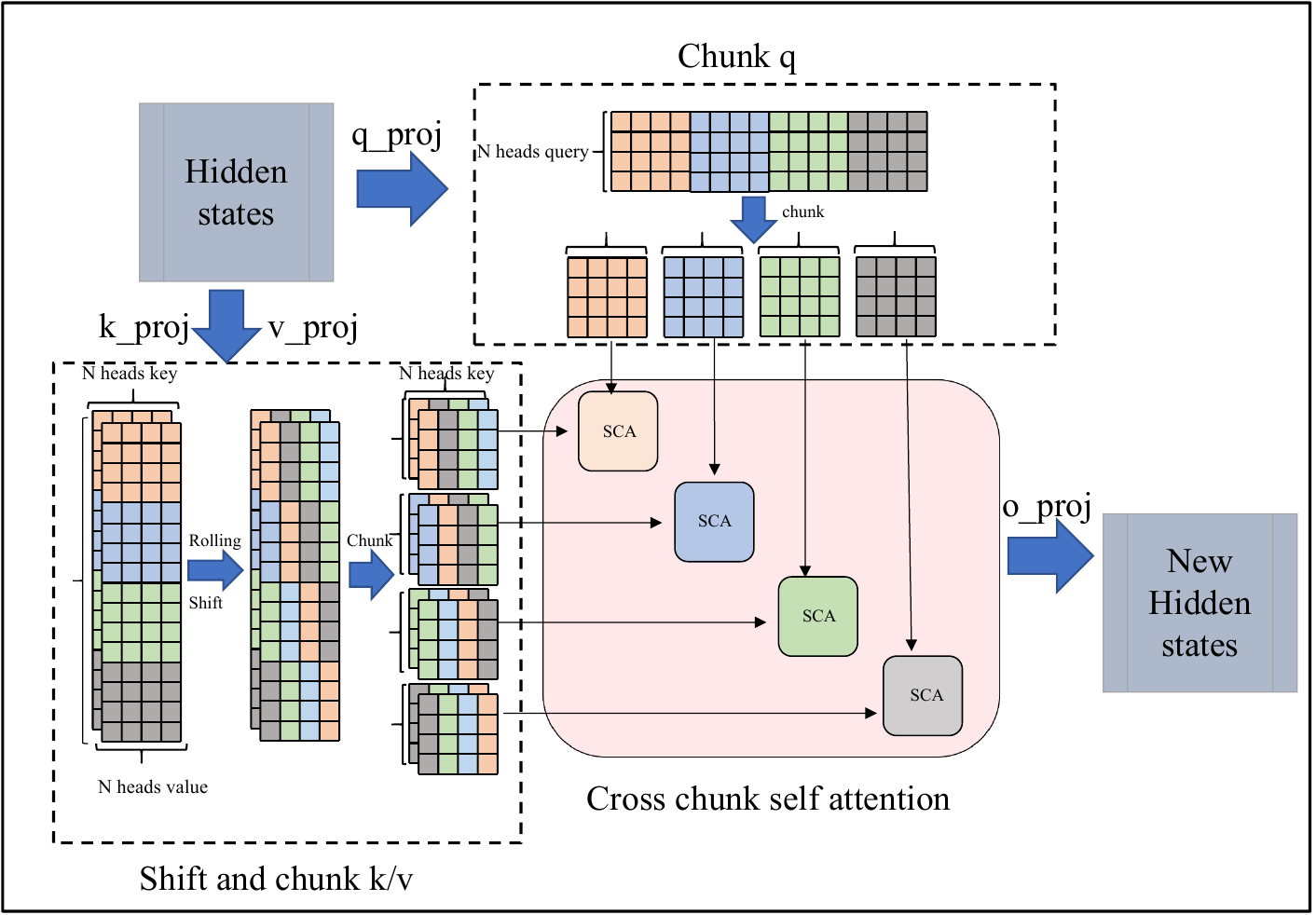}
       \label{fig:shiftall}
    }
    \caption{Two patterns in SCCA, the left figure shows the half head will be right shift half group tokens, the right figure shows the $SCCA_{flow} pattern$ which each head shift different group number tokens to make query can attend to all tokens in attention operation}
    \label{fig:shiftkv}
\end{figure*}
\section{Shifted cross chunk attention}
Standard self attention using softmax to compute attention weights of Query $Q=\{ Q_1,Q_2,...,Q_h \}$ attending Key $K= \{K_1,K_2,...,K_h \}$, then dot Value $V= \{V_1,V_2,...,V_h \}$ following equation (1), $h$ is the head number,$Q_i$ $K_i$ and $Q_i$ represents the ith head vector in multi head attention. $N$ represents the token number in a sequence. $k_i$, $v_i$ and $q_i$ represents ith token vector in one head.
\begin{equation}\small
    Attention(Q, K, V ) = softmax(\frac{QK^T}{\sqrt{d}} )V
\end{equation}
We first split QKV vector into $m$ chunks, each chunk contains $w$ tokens, where $m=\frac{N}{w}$. Different from $S^2 attention$, which redistricts window by shift $Q$, $K$ and $V$, we just shift K and V and keep the window partition still to make query $Q_{ci}$ to attend $K_{cj}$ where $1<=j<=m$. Figure \ref{fig:shiftkv} shows two different patterns in Shifted cross chunk attention (SCCA for abbreviation) in multi head attention scenario. Figure \ref{fig:shifthalf} represents the $SCCA_{fixed}$ pattern, in which half heads can only attend within window and the other heads can attend to other window by using SCCA. Figure \ref{fig:shiftall} shows the $SCCA_{flow}$ pattern, each window can attend to other windows by shifting KV in different distance in different heads. Where $g= \frac{w}{2}$
\subsection{Fixed shifted cross chunk attention}
$K_i$ and $V_i$ in ith head with shifting will be rearranged into $SK_i$ and $SV_i$ like equation (2) and equation (3)
\begin{equation}\small
    SK_i= \{k_{N-g-1},k_{N-g},...,k_{N-1},k_0,K_1,...,k_{N-g} \}
\end{equation}
\begin{equation}\small
    SV_i= \{v_{N-g-1},v_{N-g},...,v_{N-1},v_0,v_1,...,v_{N-g} \}
\end{equation}
After shifting KV vector we need split then into different chunks based on window, figure \ref{fig:shiftkv} is an example which contains four chunks, and each chunk is composed of four tokens.
Then the KV matrix can be described into Equation (4) and Equation (5).
\begin{equation}\small
    K= \{SK_1,SK_2,...,SK_{h/2},K_{h/2+1},K_{h/2+2},,...,K_{h} \}
\end{equation}
\begin{equation}\small
    V= \{SV_1,SV_2,...,SV_{h/2},V_{h/2+1},V_{h/2+2},...,V_h \}
\end{equation}
$Q_{ci}$, $K_{ci}$ and $V_{ci}$ respectively represents ith chunk in multi head Q K V vector, and each chunk contains shifted and non-shifted KV tokens. After splitting long sequence into chunks, SCCA conduct attention operation within each chunk following Equation (6).
\begin{equation}\tiny
   Attention(Q, K, V )= \begin{bmatrix}
 softmax(\frac{Q_{c1}K_{c1}^T}{\sqrt{d}})V_{c1} \\ \\
 softmax(\frac{Q_{c2}K_{c2}^T}{\sqrt{d}})V_{c2}\\ \\
 ...\\ \\
softmax(\frac{Q_{c3}K_{c3}^T}{\sqrt{d}})V_{c3} \\ 
\end{bmatrix}
\end{equation}
\subsection{Flow shifted cross chunk attention}
Different like lase section we just shift fix half group size, this section we propose a new shift pattern which different head shift different chunk size to explore the receptive field in one layer. 

Figure \ref{fig:shiftall} shows the certain process in $SCCA_{flow}$ during shifting all heads in a different shift distance. In this situation, shift pattern follows the group number. The target of this pattern is to simulate the receptive field of full attention through multi head mechanism.Algorithm \ref{tab:shiftcode} shows the implementation pseudocode in $SCCA_{flow}$. Shifting KV vector and keep query still as Equation (7) (8) and Equation (9) can explore respective field in one attention layer by accumulating computing results of multiple heads. Where $w$ represents the group size in each chunk, and $m=\frac{N}{w}$, which means one sequence can be split into $m$ chunks. $t= \frac{h}/{m}$ represents the head number which have the same shift distance.

$Q_{i_j}=\{ q_{jw+1}, q_{iw+1}, q_{iw+1}, ... , q_{(j+1)w}\}$ means jth chunk query vector in head $i$, $K_{i_j}=\{ k_{jw+1}, k_{iw+1}, k_{iw+1}, ... , k_{(j+1)w}\}$ means jth chunk key vector in head $i$, $V_{i_j}=\{ v_{jw+1}, v_{iw+1}, v_{iw+1}, ... , v_{(j+1)w}\}$ means jth chunk value vector in head $i$.
\begin{equation}\tiny
    Q= \begin{bmatrix}
  Q_{1_{1}},Q_{1_{2}},Q_{1_{3}},...,Q_{{1}_{m}} \\ 
  Q_{2_{1}},Q_{2_{2}},Q_{2_{3}},...,Q_{{2}_{m}} \\
  ... \\
  Q_{h_{1}},Q_{h_{2}},Q_{h_{3}},...,Q_{{h}_{m}}\\
\end{bmatrix} 
\end{equation}
\begin{equation}\tiny
    K= \begin{bmatrix}
  K_{1_{1}},K_{1_{2}},...,K_{1_{m-1}},K_{1_{m}} \\ 
  ...\\
  K_{t_{1}},K_{t_{c2}},...,K_{t_{cm-1}},K_{t_{m}} \\
  K_{{t+1}_{2}},K_{{t+1}_{3}},...,K_{{t+1}_{m}}, K_{{t+1}_{1}} \\
  ... \\
  K_{{2t}_{2}},K_{{2t}_{3}},...,K_{{2t}_{m}}, K_{{2t}_{1}} \\
  ...\\
  ...\\
  K_{{h-t+1}_{m}},K_{{h-t+1}_{1}},...,K_{{h-t+1}_{m-2}},K_{{h-t+1}_{m-1}} \\
  ... \\
  K_{{h}_{m}},K_{h_{1}},...,K_{{h}_{m-2}},K_{h_{m-1}} \\
\end{bmatrix} 
\end{equation}
\begin{equation}\tiny
    V= \begin{bmatrix}
  V_{1_{1}},V_{1_{2}},...,V_{1_{m-1}},V_{1_{m}} \\ 
  ...\\
  V_{t_{1}},V_{t_{c2}},...,V_{t_{cm-1}},V_{t_{m}} \\
  V_{{t+1}_{2}},V_{{t+1}_{3}},...,V_{{t+1}_{m}}, V_{{t+1}_{1}} \\
  ... \\
  V_{{2t}_{2}},V_{{2t}_{3}},...,V_{{2t}_{m}}, V_{{2t}_{1}} \\
  ...\\
  ...\\
  V_{{h-t+1}_{m}},V_{{h-t+1}_{1}},...,V_{{h-t+1}_{m-2}},V_{{h-t+1}_{m-1}} \\
  ... \\
  V_{{h}_{m}},V_{h_{1}},...,V_{{h}_{m-2}},V_{h_{m-1}} \\
\end{bmatrix} 
\end{equation}

After shifting we split into sequencs and conduct window attention like Equation (6) to reduce computing memory and time cost.
\section{LongMixed}
\begin{figure}
    \centering
    \subfigure[Dilated distance=2]{
     \includegraphics[width=0.2\textwidth]{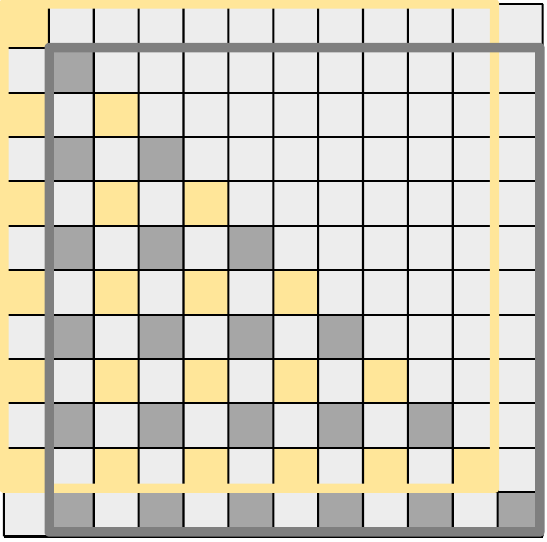}
     \label{fig:dilate2}
    }
    \subfigure[Dilated distance=4]{
       \includegraphics[width=0.2\textwidth ]{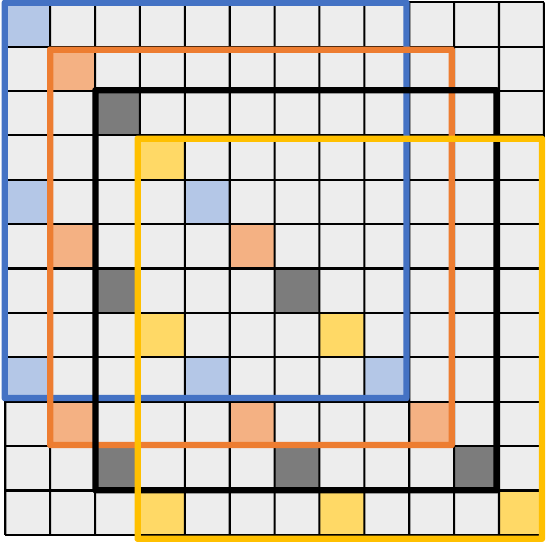}
       \label{fig:dilate4}
    }
    \caption{Illustration of two different pattern in DAT, DAT conduct sliding window attention in each head. Top figure shows the pattern dilated distance=2, and adjacent head tokens start from 1 and 2 respectively, then repeat this process $\frac{h}{2}$ times. Bottom figure shows the pattern dilated distance=4, and adjacent head tokens start from 1, 2, 3, 4 respectively, then repeat this process $\frac{h}{4}$ times in multi head attention process}
    \label{fig:dilate}
\end{figure}
In this section we propose a new combination that different sparse attention can be combined to improve model performance in fine-tuning process to extrapolate context length in LLMs. 

Inspired by DAT \cite{hassani2023dilated} and LongNet \cite{ding2023longnet}, we propose Shifted Dilated Attention (SDA), a sparse global attention. Figure \ref{fig:dilate} shows the two patterns of SDA. Similar to DAT, we select computing tokens in global space, different from DAT conducting shifted computing in different attention layer, we shifting start position in each head, and this operation is similar to Dilated Attention (DA) in LongNet. The difference from LongNet is that DA select dilated tokens in a segment which contains a subset of global tokens, and we directly conduct DA in the whole global space and do not split any segments or chunks. Figure \ref{fig:dilate2} shows the SDA attention pattern where dilated distance equal to 2 and \ref{fig:dilate4} is the pattern where dilated distance equal to 4. This method conduct a sliding dilated token selection in a sequence in different heads, and start index is begin from $1, 2, 3,...,\theta$, where $\theta$ is dilated distance. 

Different attention pattern can be combined during fine-tuning process to extrapolate context length. In this section, we combine $SCCA_{fixed}$ and SDA into LongMixed.
\begin{table}
\centering 
\caption{Perplexity of models extended to 8k context size via PI and different sparse attention pattern on PG19 validation set}
\begin{tabular}{ll}
\hline
\textbf{attention} &8192\\
\hline
$LLaMA2$ &1000 \\
$S^2$ \cite{chen2023longlora} 9.41 \\
$SCCA_{fixed}$ &9.17 \\  
$SCCA_{flow}$ &9.47\\  
$LongMixed$ &8.73 \\ \hline
\end{tabular}
\label{tab:pg19}
\end{table}
\section{Experiments}
\subsection{Settings}
\begin{table*}
\centering 
\caption{Perplexity of models extended to 8k context size via PI and different sparse attention pattern on PG19 validation set and Proof test set. The training dataset come from a subset of RedPajama .We show that our proposed attention pattern have a better performance in 8k context than $S^2$ attention}
\begin{tabular}{lllll}
\hline
\textbf{PG19}\\ \hline
\textbf{attention} & 1024&2048&4096&8192\\
\hline
$S^2$ \cite{chen2023longlora} & 11.71&10.73&9.98&9.41 \\
$SCCA_{fixed}$ &11.26&10.33&9.63&9.17 \\  
$SCCA_{flow}$ &11.59&10.64&9.94&9.47\\  
$LongMixed$ &10.49 &9.65&9.10&8.73 \\ \hline
\end{tabular}
\begin{tabular}{llll}
\hline
\textbf{Proof}\\ \hline
 1024&2048&4096&8192\\
\hline
 3.99&3.83&3.15&2.96\\
 3.95&3.43&3.09&2.88 \\ 
3.99&3.47&3.13&2.91 \\  
3.96 &3.46&3.12&2.90 \\ \hline
\end{tabular}
\label{tab:ppl}
\end{table*}
{\bf Datasets}, we use a subset of RedPajama \cite{Redpajama} dataset for fine-tuning next token prediction task, we select training samples which token length larger than 8192 by using LLaMA tokenizer. The number of total training samples is 21768. We evaluate the perplexity on PG19 validation split and Proof-pile dataset \cite{proofpile} test split. 

{\bf Model } We select LLaMA2-7B base model as our evaluation model and compare to the most similar attention pattern $S^2$ attention. Both two attention pattern conduct the same training settings.

{\bf Attention pattern setting} For $SCCA_{fixed}$ and $SCCA_{flow}$, we set chunk number $m=4$ and $SCCA_{fixed}$ right shift half $\frac{N}{m}$ tokens in half heads, $SCCA_{flow}$ shift $iw$ tokens in different head. 
For LogMixted, 8 heads are selected to conduct $SDA_2$ and 16 heads are selected to conduct $SDA_4$, the other heads conduct $SCCA_{fixted}$.

{\bf Training and evualtion setting} We use DeepSpeed \cite{deepspeed} in Stage 3 during fine-tuning and LoRA\cite{hu2022lora} setting is the same as Longlora \cite{chen2023longlora}. We use Adamw Optimizer and the learning rate is set to 2e-5, we use constant and linear learning rate with warmup, warmup step is 20. We set per-device batch size as 1 in 32G 8*V100, which means the global batch size is 8. We fine-tune 1 epoch in 21768 training samples in RedPajama. We evaluation perplexity scores at various evaluation context window sizes, ranging from 1024 to 8192. For evaluation efficiency, we set the stride of the sliding window to 256 and use 4-bit quantization technique.
\subsection{language modeling results}
In Table \ref{tab:ppl}, we report the perplexity for our models and baseline $S^2$ attention on Proof-pile and PG19 datasets. Under certain training context lengths, $SCCA_{fixed}$ and LongMixed achieve better perplexity with 1024,2048,4096 even in 8192 context than $S^2$ attention. This indicates the effectiveness of our efficient attention pattern. In Table \ref{tab:ppl}, for the same training and evaluation context length cases, the perplexity decreases as the context size increases. we find some perplexity degradation on small context sizes for the extended models. This is a known limitation of Position Interpolation.





\appendix

\section{Example Appendix}
\label{sec:appendix}

To be continued.

\end{document}